\documentclass{article}

\usepackage{booktabs}       %
\usepackage{amsfonts,amssymb,mathtools}       %

\usepackage{wrapfig}

\usepackage{natbib} 
\usepackage{floatrow}
\usepackage{subcaption}
\usepackage{multirow}
\usepackage{enumitem}

\usepackage{flushend}
\usepackage{hyperref}
\usepackage{cleveref}
\usepackage{acronym}
\usepackage{color,xcolor}
\usepackage{verbatim}
\usepackage{siunitx}
\usepackage{array}
\newcolumntype{P}[1]{>{\centering\arraybackslash}p{#1}}
\usepackage{rotating}
\usepackage{floatrow}
\newfloatcommand{capbtabbox}{table}[][0.85\FBwidth]

\crefname{appsec}{Appendix}{Appendices}

\newcommand\bbR{\ensuremath{\mathbb{R}}} %
\newcommand\C{\ensuremath{\mathbf{C}}} %

\DeclareMathOperator*{\argmax}{arg\,max}
\newcommand\Nml{\mathcal{N}}

\usepackage{todonotes}
\usepackage{soul}

\usepackage[utf8]{inputenc}
\RequirePackage{luatex85}
\usepackage{pgfplots}
\pgfplotsset{compat=newest}
\pgfplotsset{every axis legend/.append style={%
cells={anchor=west}}
}
\usepgfplotslibrary{polar}
\usetikzlibrary{arrows}
\tikzset{>=stealth'}

\definecolor{C1}{rgb}{0.0, 0.447, 0.741}
\definecolor{C1_light}{rgb}{0.0, 0.6032388663967612, 1.0}
\definecolor{C2}{rgb}{0.85, 0.325, 0.098}
\definecolor{C3}{rgb}{0.929, 0.694, 0.125}
\definecolor{C4}{rgb}{0.494, 0.184, 0.556}
\definecolor{C5}{rgb}{0.466, 0.674, 0.188}
\definecolor{C6}{rgb}{0.301, 0.745, 0.933}
\definecolor{C7}{rgb}{0.635, 0.078, 0.184}

\definecolor{nice-red}{HTML}{E41A1C}
\definecolor{nice-orange}{HTML}{FF7F00}
\definecolor{nice-yellow}{HTML}{FFC020}
\definecolor{nice-green}{HTML}{4DAF4A}
\definecolor{nice-blue}{HTML}{377EB8}
\definecolor{nice-nice-red}{HTML}{984EA3}
\usepgfplotslibrary{groupplots}

\usetikzlibrary{shapes.geometric, arrows}

\tikzstyle{startstop} = [rectangle, rounded corners, minimum width=2cm, minimum height=1cm,text centered, draw=black, fill=none]
\tikzstyle{arrow} = [thick,->,>=stealth]

\usetikzlibrary{backgrounds}

\usepackage{ifthen}

\newcommand{\ML}{\text{ML}}

\usepackage{microtype}
\usepackage{graphicx}
\usepackage{booktabs} %

\usepackage{hyperref}

\usepackage[accepted]{icml2020}

\icmltitlerunning{Scalable Identification of Partially Observed Systems \\ with Certainty-Equivalent EM}

\def\figtocap{-2em}

\begin{document}

\twocolumn[
\icmltitlerunning{Scalable Identification of Partially Observed Systems with CE-EM}
\icmltitle{Scalable Identification of Partially Observed Systems with Certainty-Equivalent EM}

\icmlsetsymbol{equal}{*}

\begin{icmlauthorlist}
\icmlauthor{Kunal Menda}{equal,st}
\icmlauthor{Jean de Becdelièvre}{equal,st}
\icmlauthor{Jayesh K. Gupta}{equal,st}\\
\icmlauthor{Ilan Kroo}{st}
\icmlauthor{Mykel J. Kochenderfer}{st}
\icmlauthor{Zachary Manchester}{st}
\end{icmlauthorlist}

\icmlaffiliation{st}{Stanford University, CA, USA}

\icmlcorrespondingauthor{Kunal Menda}{kmenda@stanford.edu}
\icmlcorrespondingauthor{Jean de Becdelièvre}{jeandb@stanford.edu}
\icmlcorrespondingauthor{Jayesh K. Gupta}{jkg@cs.stanford.edu}

\icmlkeywords{System Identification, Expectation-Maximization, Robotics}

\vskip 0.3in
]

\printAffiliationsAndNotice{\icmlEqualContribution} %

\begin{abstract}
  System identification is a key step for model-based control, estimator design, and output prediction.
  This work considers the offline identification of partially observed nonlinear systems.
We empirically show that the certainty-equivalent approximation to expectation-maximization can be a reliable and scalable approach for high-dimensional deterministic systems, which are common in robotics. We formulate certainty-equivalent expectation-maximization as block coordinate-ascent, and provide an efficient implementation.
  The algorithm is tested on a simulated system of coupled Lorenz attractors, demonstrating its ability to identify high-dimensional systems that can be intractable for particle-based approaches. 
  Our approach is also used to identify the dynamics of an aerobatic helicopter. 
  By augmenting the state with unobserved fluid states, a model is learned that predicts the acceleration of the helicopter better than state-of-the-art approaches.
  The codebase for this work is available at \url{https://github.com/sisl/CEEM}.
\end{abstract}

\section{Introduction}
\label{sec:introduction}

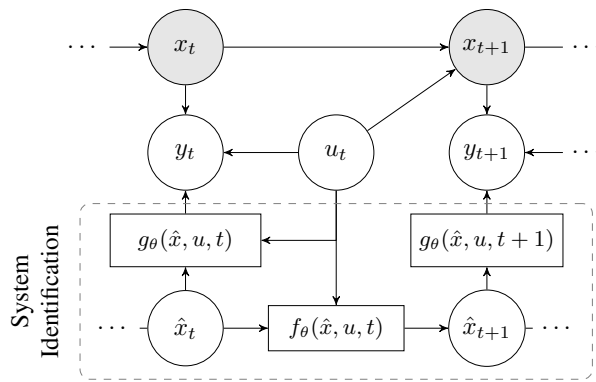
\begin{figure}[t]
    \centering
    \scalebox{0.95}{\begin{tikzpicture}
        
    \def\horsep{12em};
    \def\versep{7em};
    
    \def\circrad{3em};

    \def\Pheight{\circrad*0.7};

	\node[circle, minimum size=\circrad, draw=black, fill=gray!20] (xt) at (0,0) {$x_t$};
	
	\node[circle, minimum size=\circrad, draw=black, fill=gray!20] (xtp1) at ($(xt)+(\horsep,0)$) {$x_{t+1}$};

    \node[circle, minimum size=\circrad, draw=black, fill=white] (ut) at ($(xt)!0.5!(xtp1) + (0,-0.6*\versep)$) {$u_t$};

    \node[circle, minimum size=\circrad, draw=black, fill=white] (ot) at ($(xt) + (0,-0.6*\versep)$) {$y_t$};

    \node[circle, minimum size=\circrad, draw=black, fill=white] (otp1) at ($(xtp1) + (0,-0.6*\versep)$) {$y_{t+1}$};

    \draw[->] (xt) -- (xtp1);

    \draw[->] (ut) -- (xtp1);

    \draw[->] (ut) -- (ot);

    \draw[->] (xt) -- (ot);

    \draw[->] (xtp1) -- (otp1);

    \node[left of=xt, xshift=-0.1*\horsep] (cdots1) {$\cdots$};
    \draw[->] (cdots1) -- (xt);
    
    \node[right of=otp1, xshift=0.1*\horsep] (cdotso) {$\cdots$};
    \draw[<-] (otp1) -- (cdotso);
    
    \node[right of=xtp1, xshift=0.1*\horsep] (cdots2) {$\cdots$};
    \draw[-] (xtp1) -- (cdots2);

    \node[circle, minimum size=\circrad, draw=black, fill=white] (bxt) at ($(ot)+(0,-\versep)$) {$\hat{x}_t$};
	
	\node[circle, minimum size=\circrad, draw=black, fill=white] (bxtp1) at ($(bxt)+(\horsep,0)$) {$\hat{x}_{t+1}$};

    \node[rectangle, minimum width=\circrad*1.8, minimum height=\circrad*0.6, draw=black, fill=white] (f1) at ($(bxt)!0.5!(bxtp1)$) {\footnotesize$f_\theta(\hat{x},u, t)$};

    \node[rectangle, minimum width=\circrad*2, minimum height=\Pheight, draw=black, fill=white] (P1h) at ($(bxt)!0.5!(ot)$) {\footnotesize$g_\theta(\hat{x}, u, t)$};
    
    \node[rectangle, minimum width=\circrad*2, minimum height=\Pheight, draw=black, fill=white] (P2h) at ($(bxtp1)!0.5!(otp1)$) {\footnotesize$g_\theta(\hat{x}, u,t+1)$};

    \draw[->] (ut.south) |- (P1h.east);
    \draw[->] (bxt) -- (f1);
    \draw[->] (f1) -- (bxtp1);

    \draw[->] (bxt) -- (P1h);
    \draw[->] (P1h) -- (ot);
    \draw[->] (bxtp1) -- (P2h);
    \draw[->] (P2h) -- (otp1);
    
    \node[left of=bxt] (cdots1) {$\cdots$};
    \draw[-] (cdots1) -- (bxt);
    
    \node[right of=bxtp1] (cdots2) {$\cdots$};
    \draw[-] (bxtp1) -- (cdots2);

    \draw[->] (ut) -- (f1);

    \coordinate (Phm) at ($(P1h)!0.5!(P2h)$);
    \coordinate (bxm) at ($(bxt)!0.5!(bxtp1)$);
    
    \node[dashed, rounded corners, rectangle, minimum height=\versep, minimum width=1.7*\horsep, draw=black!50, anchor=center] (box2) at ($(Phm)!0.5!(bxm) - (0,0.1)$) {};
    
    \node[rotate=90, align=center] (inflearn) at ($(box2)+(-\horsep,0)$) {System\\Identification};
	
\end{tikzpicture}}
    \vspace{-0.75em}
    \caption{A graphical model representing a partially observed dynamical system. Gray-box identification algorithms attempt to search a model class of dynamics and observation models for the model that maximizes the likelihood of the observations.}
    \label{fig:graphicalmodel}
    
\end{figure}

The performance of controllers and state-estimators for non-linear systems depends heavily on the quality of the model of system dynamics~\citep{hou2013model}. 
System-identification addresses the problem of learning or calibrating dynamics models from data~\citep{ljung1999system}, 
which is often a time-history of observations of the system and control inputs. 
In this work, we address the problem of learning dynamics models of partially observed systems (shown in \Cref{fig:graphicalmodel}) that are high-dimensional and non-linear.
We consider situations in which the system's state cannot be inferred from a single observation, but instead requires inference over time-series of observations. 

\label{appsec:probstate}

The problem of identifying systems from partial observations arises in robotics~\citep{punjani2015deep, cory2008experiments, ordonez2017learning} as well as domains such as chemistry \cite{gustavsson_survey_1975} and biology~\cite{sun2008ekfbio}. 
In many robotic settings, we have direct measurements of a robot's pose and velocity, but in many cases we cannot directly observe relevant quantities such as the temperature of actuators, the state the environment around the robot, or the intentions of other agents. 
For example, \citet{abbeel2010heli} attempted to map the pose and velocity of an aerobatic helicopter to its acceleration. They found their model to be inaccurate when predicting aggressive maneuvers because of the substantial airflow generated by the helicopter that affected the dynamics.
Since it is often impossible to directly measure the state of the airflow around a vehicle, identification must be with only partial observability.

System identification is a mature field with a rich history~\citep{ljung1999system,ljung2010perspectives}. 
A variety of techniques have been proposed to learn predictive models from time series data. 
Autoregressive approaches directly map a time-history of past inputs to observations, without explicitly reasoning about unobserved states ~\citep{billings2013narmax}, and are the state-of-the-art approach to the aforementioned problem of modeling the aerobatic helicopter~\cite{punjani2015deep}.
In contrast, state-space models (SSM) assume an unobserved state $x_t$ that evolves over time and \textit{emits} observations $y_t$ that we measure, as shown in \Cref{fig:graphicalmodel}. 
Recurrent Neural Networks (RNNs)~\citep{bailer1998recurrent,zimmermann2000modeling} are a form of \textit{black-box} non-linear SSM that can be fit to observation and input time-series, and Subspace Identification (SID) 
methods~\citep{van2012subspace} can be used to fit linear SSMs.

However, in many cases prior knowledge can be used to specify structured, parametric models of the system~\citep{gupta2019laglearn,Gupta2020} in state-space form, commonly refered to as \textit{gray-box models}.
Such models can be trained with less data and used with a wider array of control and state-estimation techniques than black-box models ~\citep{gupta2019laglearn,Gupta2020,lutter2018deep,lutter2019deep}.

To identify partially observed gray-box models, the unobserved state-trajectory is often considered as a missing data, and techniques based on Expectation-Maximization (EM) are used~\citep{dempster1977em, schon201pfem, kantas2015particlesurvey,ghahramani1999learning}.
The \textit{smoothing} step (E-step) deals with state inference---the current system dynamics estimate is used to infer the distribution of unobserved state-trajectories conditioned on the observations $p(x_{1:T} \mid y_{1:T})$. 
This distribution is sometimes called joint smoothing distribution in the literature, and it is used to estimate the expected log-likelihood of the observations.
In the \textit{learning} step (M-step), the system's dynamics estimate is updated such that it maximizes the expected log-likelihood. 
The \textit{smoothing} step can typically be approached with particle approximations of $p(x_{1:T} \mid y_{1:T})$, but naive implementations of particle smoothers can be computationally intensive and become rapidly intractable in high dimensions. Across various fields and disciplines, numerous methods have been developed to alleviate the computational burden, several of which are discussed in \Cref{subsec:relatedwork}.

This work is motivated by robotics applications, in which the following assumptions are often valid:
\begin{itemize}[itemsep=2pt,nolistsep]
    \item Systems evolve nearly deterministically, which implies that the process noise is small and unimodal, and,
    \item The distribution of states conditioned on the observations $p(x_{1:T} \mid y_{1:T})$ is unimodal.
\end{itemize}
We study the benefits of making the \textit{certainty-equivalent} approximation in the E-step of the EM procedure, and we refer to this approach as \textit{CE-EM}.
Specifically, we use non-linear programming to tractably find the maximum-likelihood (ML) point-estimate of the unobserved states, and use it in lieu of a fully characterized approximate distribution. 
The contributions of this paper are to describe an efficient implementation of CE-EM, and to test the approximation against state-of-the-art approaches on a variety of system-identification problems.
We demonstrate on a system of Lorenz attractors that:
\begin{itemize}[itemsep=2pt,nolistsep]
\item CE-EM can be faster and more reliable than approaches using particle approximations,
\item CE-EM scales to high-dimensional problems, and,
\item CE-EM learns unbiased parameter estimates on deterministic systems with unimodal $p(x_{1:T} \mid y_{1:T})$.
\end{itemize}
We also demonstrate the algorithm on the problem of identifying the dynamics of an aerobatic helicopter. 
We show that a non-linear SSM can be trained with CE-EM that outperforms various approaches including the most recent work done on this dataset~\citep{punjani2015deep}. 
A codebase implementing CE-EM and other supplementary material can be found at our website: {\url{https://sites.google.com/stanford.edu/ceem/}}.

\section{Background}
\label{sec:background}
This section states the nonlinear system identification problem with partial observations and discusses approaches that use EM to solve it.

\subsection{Formal Problem Statement}

\label{sec:statement}

In this work, we assume that we are given a batch of trajectories containing observations $y_{1:T}\in \bbR^{m\times T}$ of a dynamical system as it evolves over a time horizon $T$, possibly forced by some known input sequence $u_{1:T}$.
We assume that this dynamical system has a state $x \in \bbR^n$ that evolves and generates observations according to the following equations,
\begin{equation}
\label{eq:sys}
\begin{aligned}
    x_{t+1} &= f(x_t, u_t, t) + w_t, &w_t \sim p_w(\cdot) \\ 
    y_t &= g(x_t,u_t,t) + v_t, &v_t \sim p_v(\cdot)\\
\end{aligned}
\end{equation}
where $w_t$ is referred to as the \textit{process noise} and $v_t$ as the \textit{observation noise}. Both $w_t$ and $v_t$ are assumed to be additive for notational simplicity, but this is not a required assumption.%
\footnote{In order to substitute $w_t$ or $v_t$ in the arguments of $p_v$ and $p_w$, the only requirement is to be able to express them as functions of the other terms. Hence, the formulation is amenable to any method of injecting noise into the system, so long as we can estimate the probability of the noise given the other terms in their respective equations.}
Without loss of generality, we can drop the dependence on $u_t$, absorbing it into the dependence on $t$.

We further assume that we are provided a class of parameterized models $f_\theta(x,t)$ and  $g_\theta(x,t)$ for $\theta \in \Theta$ that approximate the dynamical system's evolution and observation processes. 
The goal of our algorithm is to find the parameters $\theta$ that maximize the likelihood of the observations.
That is, we seek to find:
\begin{equation}
    \label{eqn:mlobj}
    \begin{aligned}
    \theta_{\ML} &= \underset{\theta}{\argmax}~ p(y_{1:T} \mid \theta) \\
         &= \underset{\theta}{\argmax}~  \int p(y_{1:T}, x_{1:T} \mid \theta) dx_{1:T} \\
        &= \underset{\theta}{\argmax}~  \int p(y_{1:T}\mid x_{1:T}, \theta) p(x_{1:T} \mid \theta) dx_{1:T}\\
    \end{aligned}
\end{equation}
Using the graphical model shown in \Cref{fig:graphicalmodel}, we can rewrite this integral as:
\begin{equation}
    \label{eqn:mlobj_pieces}
    \begin{aligned}
    \theta_{\ML}
        =  \underset{\theta}{\argmax}~  \int &\prod_{t=1}^T p(y_{t}\mid x_{t}, \theta) \cdot \\
        &\prod_{t=1}^{T-1} p(x_{t+1} \mid x_{t}, \theta) p(x_1 \mid \theta) dx_{1:T}\\
    \end{aligned}
\end{equation}
Finally, using the notation chosen to describe a dynamical system in \Cref{eq:sys}, we obtain:
\begin{equation}
    \label{eqn:mlobj_exp}
    \begin{aligned}
    \theta_{\ML} &= \underset{\theta}{\argmax}\, \int~\left( \prod_{t=1}^T p_v(y_t-g_\theta(x_t,t))  \right) \cdot \\ &\hspace{40pt} \left( p(x_1)\prod_{t=1}^{T-1}p_w(x_{t+1}-f_\theta(x_t,t)) \right)dx_{1:T} 
    \end{aligned}
\end{equation}
The Expectation-Maximization (EM) algorithm has been used in the literature to solve problems of this form.

\subsection{Expectation-Maximization for SSM Identification}

The EM algorithm is a two-step procedure that copes with the missing state information by forming an approximation $Q(\theta, \theta_k)$ of the joint state and observation likelihood $p_{\theta}(x_{1:T}, y_{1:T})$ at the $k$th iteration of the algorithm:
\begin{equation}
    \label{eqn:Q}
    \begin{aligned}
    Q(\theta, \theta_k) = \int \log ~&p(x_{1:T}, y_{1:T}\mid \theta) \cdot 
    \\
    &p(x_{1:T} \mid y_{1:T}, \theta_k)dx_{1:T}
    \end{aligned}
\end{equation}
Classical proofs show that maximizing $Q(\theta, \theta^k)$ with respect to $\theta$ results in increasing $p_{\theta}(x_{1:T}, y_{1:T})$ ~\citep{dempster1977em}. The EM algorithm iteratively performs two steps:
\begin{enumerate}[itemsep=2pt,nolistsep]
    \item Compute $Q(\theta, \theta_k)$
    \item Update $\theta_{k+1} = \underset{\theta}{\argmax} \; Q(\theta, \theta_k)$
\end{enumerate}
The evaluation of $Q(\theta, \theta_k)$ in the E-step requires the estimation the $p(x_{1:T} \mid y_{1:T}, \theta_k)$, as well as the integration of the likelihood over this distribution. In the linear Gaussian case, EM can be performed exactly using Kalman smoothing~\citep{rauch1965maximum, ghahramani1996ldsem}. In the more general case, there is no analytic solution, and previous work on approximating the E-step is summarized in the next section.

\vspace{-0.5em}

\subsection{Related Work}
\label{subsec:relatedwork}

\vspace{-0.3em}

There are general approaches that are based on particle representations of the joint states-observations likelihood which use Sequential Monte-Carlo (SMC) methods such as Particle Smoothing (PS) \citep{schon201pfem, kantas2015particlesurvey}. 
With enough particles, PS can handle any system and any joint distribution of states and observations. 
However, PS suffers from the \textit{curse of dimensionality}, requiring an intractably large number of particles if the state space is high-dimensional \citep{snyder2008obstacles, kantas2015particlesurvey}. 
In the simplest form, both the E-step and the M-step can be quadratic in complexity with respect to the number of particles \citep{schon201pfem}. 
An important body of work has attempted to alleviate some of this burden by using Nested SMC~\cite{naesseth2015nested} and forward filtering-backward simulation (FFBSi), \citep{lindsten2013backward}, and conditional particle filtering \cite{linstenConditional} in the E-step.
PS can also require variance reduction techniques, such as full-adaptation, to perform reliably, and likelihood estimates in the E-step can be noisy \citep{svensson2018learning}. 
Stochastic Approximation EM can be used to stabilize learning in the presence of this noise~\cite{delyon1999saem, svensson2018psaem}.

Another group of methods is based on linearizing the dynamics around the current state estimates to obtain a time-dependant linear Gaussian dynamical system, to which we can apply Kalman smoothing techniques.
\citet{ghahramani1999learning} use Extended Kalman Smoothing (EKS) for a fast E-step and show that, when using radial basis functions to describe the system dynamics, no assumption is required in the M-step. 
The drawbacks of this approach are that the number of radial basis functions required to accurately represent a function grows exponentially with the input dimension, that the user cannot specify a gray-box SSM, and that EKS performance can vary dramatically with the hyperparameters.
\citet{goodwin2005approximate} proposed a method called MAP-EM, which linearizes around the state-trajectory that maximizes joint state and observation likelihood, approximates process and observation noise as Gaussian, and performs a \textit{local} version of EM using this linearization.
In comparison, the method we consider assumes that the maximum likelihood state-trajectory estimate concentrates all of the probability mass. \citet{goodwin2005approximate} call this  simplification Certainty Equivalent EM (CE-EM) and report worse results than their approach on simple examples.

This paper shows that CE-EM can actually be a fast, simple, and reliable approach. 
The method we use here is tailored to systems that are not dominated by process noise, in which the state-trajectory distribution is unimodal, and whose high dimensional state-space make other methods intractable.
Such scenarios are regularly encountered in robotics, but perhaps less so in other applications domains of system identification such as chemistry, biology, or finance.

\vspace{-0.5em}
\section{Methodology}
\label{sec:methodology}

This section introduces the CE-EM approximation and presents an efficient algorithm to identify high-dimensional robotic systems.

\vspace{-0.5em}

\subsection{Certainty-Equivalent Expectation Maximization}

Under the certainty-equivalence condition, the distribution of states conditioned on observations is assumed to be a Dirac delta function. 
That is, we assume:
\begin{equation}
    p(x_{1:T} | y_{1:T}, \theta) = \delta_{x_{1:T}^{\ML}(\theta)}(x_{1:T})
\end{equation}
where
\begin{equation}
    \label{eqn:x_ML_CE}
    \begin{aligned}
    x_{1:T}^{\ML}(\theta) &= \argmax_{x_{1:T}} p(x_{1:T} \mid y_{1:T}, \theta) \\
    &= \argmax_{x_{1:T}} p(x_{1:T}, y_{1:T} \mid \theta) / p(y_{1:T} \mid \theta) \\
    &= \argmax_{x_{1:T}} p(x_{1:T}, y_{1:T} \mid \theta)
    \end{aligned}
\end{equation}
In other words, this assumption implies that there is only one state trajectory that satisfies the system dynamics, and is coherent with the data.

The E-step can be rewritten as:
\begin{equation}
    \label{eqn:Q_CE}
    \begin{aligned}
    Q(\theta, \theta_k) &= \int \log~p(x_{1:T}, y_{1:T}\mid \theta)\cdot\\ &\hspace{5em}\delta_{x_{1:T}^{\ML}(\theta_k)}(x_{1:T})dx_{1:T} \\
    &= \log p(x^{\ML}_{1:T}(\theta_k), y_{1:T} \mid \theta)
    \end{aligned}
\end{equation}

By observing the maximization of the expression in \Cref{eqn:x_ML_CE} as the E-step, and maximizing the expression in \Cref{eqn:Q_CE} as the M-step, we see that CE-EM is simply block coordinate-ascent on the single joint-objective:
\begin{equation}
    \label{eqn:sislobj}
    \begin{split}
    J(x_{1:T},\theta)=& \log p(x_{1:T}, y_{1:T} \mid \theta)  \\
    =&\log p(x_1) + \sum_{t=1}^{T} \log p_v(y_t-g_\theta(x_t,t)) \\ &+ \sum_{t=1}^{T-1} \log p_w(x_{t+1}-f_\theta(x_t,t))
    \end{split}
\end{equation}%
Jointly maximizing this objective over $x_{1:T}$ and $\theta$ yields $\theta_{\ML}$. 

Though this objective can be optimized as is by a non-linear optimizer, it is not necessarily efficient to do so since $\theta$ and $x_{1:T}$ are highly coupled, leading to inefficient and potentially unstable updates.
For this reason, we still opt for the block coordinate-ascent approach similar to EM, where $\theta$ and $x_{1:T}$ are each sequentially held constant while the other is optimized.
By viewing EM as block coordinate-ascent, we may also borrow good practices typically employed when running the algorithm~\citep{shi2016primer}. 
In particular, we employ trust-region regularization in order to guarantee convergence in even non-convex settings~\citep{grippo2000convergence}.

\subsubsection{Smoothing}

At iteration $k$, we first perform \textit{smoothing} by holding $\theta$ constant and finding the ML point-estimate for $x_{1:T}$ as follows:
\begin{equation}
\label{eqn:smoothinguncon}
    x^{(k)}_{1:T} = \underset{x_{1:T}}{\argmax}~ J(x_{1:T}, \theta^{(k-1)}) - \rho_x \lVert x_{1:T} - x_{1:T}^{(k-1)}\rVert^2_2
\end{equation}
where $\rho_x$ scales a soft trust-region regularizer similar to damping terms found in Levenberg-Marquardt methods~\citep{Levenberg1944,Marquardt1963}.
We note that the Hessians of this objective with respect to $x_{1:T}$ are block-sparse, and as a result this step can be efficiently solved with a second-order optimizer in $\mathcal{O}(n^2 T)$.

If the cost function $J(x_{1:T}, \theta)$ is a sum of quadratic terms, i.e. if all stochasticity is Gaussian, then the smoothing can be solved with a Gauss-Newton method, where the Hessian matrix is approximated using first-order (Jacobian) information. 
In this case, the solution to the smoothing step is equivalent to iteratively performing EKS \citep{aravkin2017generalized}. However, EKS typically involves forward and backward passes along each trajectory, performing a Riccati-like update at each time step. 
Though square-root versions of the algorithm exist to improve its numerical stability~\cite{bierman1981sqrtks}, solving the problem with batch non-linear least-squares as opposed to iteratively performing a forward-backward method can improve stability for long time-series~\cite{van1981geneig}. 

\begin{algorithm}[tb]
\begin{algorithmic}
   \STATE {\bfseries Input:} observations $y_{1:T}$, control inputs $u_{1:T}$, stopping criterion $tol$
   \STATE {\bfseries Initialize:} model parameters $\theta$
   \STATE {\bfseries Initialize:} hidden state estimates $x_{1:T}$
   \REPEAT
   \STATE $J^k \xleftarrow{} J(x^k_{1:T}, \theta^k)$
   \STATE $x^{k+1}_{1:T} \xleftarrow{} \underset{\tilde{x}_{1:T}}{\arg\max}~ J(\tilde{x}_{1:T}, \theta^k) - \rho_x \lVert \tilde{x}_{1:T} - x^k_{1:T}\rVert^2_2$
   \STATE  $\theta^{k+1}  \xleftarrow{} \underset{\tilde{\theta}}{\arg\max}~ J(x^{k+1}_{1:T},\tilde{\theta}) - \rho_\theta \lVert \tilde{\theta} - \theta^k \rVert^2_2 + \log p(\tilde{\theta})$
   \UNTIL{$J^k - J(x^{k+1}_{1:T}, \theta^{k+1}) \leq tol$}
\end{algorithmic}
   \caption{CE-EM Implementation}
   \label{alg:ceem}
\end{algorithm}

\subsubsection{Learning}

In the \textit{learning} step, we hold $x_{1:T}$ constant and find:
\begin{equation}
    \label{eqn:learning}
    \theta^{(k)} = \underset{\theta}{\arg\max}~ J(x^{(k)}_{1:T},\theta) - \rho_\theta \lVert \theta - \theta^{(k-1)}\rVert^2_2 + \log p(\theta)
\end{equation}
Again, $\rho_\theta$ scales a soft trust-region regularizer, and specifying $\log p(\theta)$ allows us to regularize $\theta$ toward a prior. 
The above optimization problem can be solved using any non-linear optimizer.
We find the Nelder-Mead~\cite{Gao2012NelderMead} scheme well-suited for small parameter spaces, and first-order schemes such as Adam~\cite{kingma2014adam}, or quasi-second-order schemes such as L-BFGS~\cite{liu1989lbfgs} suited for larger parameter spaces such as those of neural networks. 
The routine, summarized in Algorithm \ref{alg:ceem}, iterates between the smoothing and learning steps until convergence. 

It should be noted that making the certainty-equivalent approximation is generally known to bias parameter estimates~\cite{celeux1992emclustering}, but can yield the correct solutions under the assumptions we make~\cite{neal1998emvariants}.

\section{Experiments}
\label{sec:experiments}

Readers of \citet{goodwin2005approximate} will likely be left with the incorrect impression that CE-EM is an inferior method that would perform poorly in practice.
The objective of our experiments is to demonstrate that the CE-EM algorithm is capable of identifying high-dimensional non-linear systems in partially observed settings. 
We do so in simulation by identifying the parameters of a system of partially observed coupled Lorenz attractors, as well as by identifying the dynamics of a real aerobatic helicopter. 
In the second experiment, we build on previous analysis of the dataset \citep{abbeel2010heli, punjani2015deep} by attempting to characterize the interaction of the helicopter with the fluid around it, without having any direct observation of the fluid state.
Instructions for reproducing all experiments are included in the supplementary material.

\subsection{Identification of Coupled Lorenz Systems}
\label{subsec:lorenz}
In this experiment, we show that:
\begin{enumerate}[itemsep=2pt,nolistsep]
    \item CE-EM learns unbiased parameter estimates of systems that are close to deterministic, and,
    \item CE-EM scales to high-dimensional problems in which particle-based methods can be intractable.
\end{enumerate}
To justify these claims, we use a system that is sufficiently non-linear and partially observable to make particle-based smoothing methods intractable.
We choose a system of coupled Lorenz attractors for this purpose, owing to their ability to exhibit chaotic behavior and their use in non-linear atmospheric and fluid flow models~\citep{bergs1984lorenz}.
Arbitrary increases in state dimensionality can be achieved by coupling multiple individual attractors.
The state of a system with $K$ coupled Lorenz attractors is $x\in \bbR^{3K}=\{\ldots,x_{1,k},x_{2,k},x_{3,k},\ldots\}$. 
The dynamics of the system are as follows:
\begin{equation}
    \begin{aligned}
        \dot{\tilde{x}}_{1,k} &= \sigma_k(x_{2,k}-x_{1,k})\\
        \dot{\tilde{x}}_{2,k} &= x_{1,k}(\rho_k-x_{3,k})-x_{2,k}\\
        \dot{\tilde{x}}_{3,k} &= x_{1,k} x_{2,k} - \beta_kx_{3,k}\\
        \dot{x} &= \dot{\tilde{x}} + Hx
    \end{aligned}
\end{equation}
where $H$ is an $\bbR^{3K\times 3K}$ matrix.

We nominally set the parameters $(\sigma_k, \rho_k, \beta_k)$ to the values $(10,28,8/3)$, and randomly sample the entries of $H$ from a normal distribution to generate chaotic and coupled behavior between attractors, while avoiding self-coupling. 
These parameters are estimated during identification.
In order to make the system partially observed, the observation $y\in \bbR^{(3K-2)}$ is derived from $x$ as follows:
\begin{equation}
    y=Cx + v,\quad v\sim \mathcal{N}(0,\sigma_v^2\mathbf{I})
\end{equation}
where $C\in\bbR^{(3K-2)\times 3K}$ is a known matrix with full row-rank, and $v$ is the observation noise sampled from a Gaussian with diagonal covariance $\sigma_v^2 \mathbf{I}$. The entries of $C$ are also randomly sampled from a standard normal distribution.
In the following experiments, we simulate the system for $T=128$ timesteps at a sample rate of $\Delta t = 0.04 \si{\second}$, and integrate the system using a $4$th-order Runge-Kutta method.
Initial conditions for each trajectory are sampled such that $x_{1,k}\sim\Nml(-6,2.5^2),x_{2,k}\sim\Nml(-6,2.5^2), x_{3,k}\sim\Nml(24,2.5^2)$.

\subsubsection{Unbiased Estimation in Deterministic Settings}
\label{subsec:ubiased}
\newcommand{\outside}[1]{\textcolor{red}{\textbf{#1}}}

\begin{table}
\centering
\caption{Mean parameter estimates and standard errors for a single Lorenz system simulated with various $\sigma_w$ and $\sigma_v$.}\label{tab:lorenz}
\resizebox{\columnwidth}{!}{\begin{tabular}{cc||lll}
\toprule
$\sigma_w$ & $\sigma_v$                            & \multicolumn{1}{c}{$\sigma$}       & \multicolumn{1}{c}{$\rho$}         & \multicolumn{1}{c}{$\beta$}       \\ \midrule
$0.001$ & $0.01$ & $10.011 (0.012)$ & $28.000 (0.001)$ & $2.667 (0.000)$ \\
$0.010$ & $0.01$ & $10.017 (0.012)$ & $28.000 (0.001)$ & $2.668 (0.001)$ \\
$0.100$ & $0.01$ & $10.064 (0.036)$ & $27.996 (0.013)$ & \outside{2.676 (0.004)} \\
$0.001$ & $0.05$ & $10.006 (0.016)$ & $27.998 (0.002)$ & $2.666 (0.001)$ \\
$0.001$ & $0.10$ & $9.998 (0.022)$ & $27.995 (0.004)$ & $2.665 (0.001)$ \\ \bottomrule
\end{tabular}}
\end{table}

To test the conditions under which CE-EM learns unbiased parameter estimates, we simulate a single Lorenz system with  $H=\mathbf{0}$ and known $C\in \bbR^{2\times3}$. 
We introduce and vary the process noise $w\sim \Nml(0, \sigma_w^2 \mathbf{I})$, and vary the observation noise coefficient $\sigma_v$, and then attempt to estimate the parameters $(\sigma, \rho, \beta)$.
Using initial guesses within $10\%$ of the system's true parameter values, we run CE-EM on a single sampled trajectory.
For each choice of $\sigma_w$ and $\sigma_v$, we repeat this process for $10$ random seeds. 

\Cref{tab:lorenz} shows the mean and standard errors of parameter estimates for various $\sigma_w$ and $\sigma_v$. 
We highlight in \textcolor{red}{red} the mean estimates that are not within two standard errors of their true value. 
We see that $\sigma$ and $\rho$ are estimated without bias for all scenarios. 
However, the estimate of $\beta$ appears to become biased as the process noise is increased, but not as the observation noise is increased. 
This supports the assumption that the objective used in CE-EM is sound when systems evolve close to deterministically, but can be biased if it is not.

\subsubsection{Comparison to Particle Based Methods}
\label{subsec:ps}
In \Cref{sec:background}, we discussed methods for parameter estimation in state-space systems that are based on particle-filtering and smoothing~\cite{kantas2015particlesurvey}. 
Since in their E-step, these methods approximate the distribution over unobserved state-trajectories as opposed to only their point estimate, such methods can be asymptotically unbiased.
However, for a finite number of particles, such methods can result in high-variance estimates. 
In this experiment, we compare the bias resulting from using CE-EM with the variance of using a state-of-the-art Particle EM algorithm.

We attempt to identify the parameters $(\sigma, \rho, \beta)$ of the same single Lorenz system as in \Cref{subsec:ubiased}. 
However, we introduce process noise $w\sim \Nml(0, 0.1^2 \mathbf{I})$ and observation noise $v\sim \Nml(0, 0.5^2 \mathbf{I})$.
We use a training dataset of four trajectories sampled with conditions specified in \Cref{subsec:lorenz}.

The performance of Particle EM can vary substantially depending on implementation of the particle filter and smoother in the E-step, and implementation of the M-step.
We use a fully-adapted particle filter with systematic and adaptive resampling, following the recommendations of \citet{doucet2000sequential, hol2006resampling}. 
Furthermore, we use the FFBSi algorithm \cite{godsill2004ffbsi, lindsten2013backward} in order to generate iid samples of smoothed state-trajectories, while avoiding complexity that is quadratic in the number of particles experienced by the forward filter-backward smoother (FFBSm) approach \cite{doucet2000sequential}.
We then use Stochastic Approximation EM (SAEM) \cite{delyon1999saem} to perform the M-step.\footnote{We have found that using FFBSi with SAEM performs more reliably than FFBSm with the M-step recommended by \cite{schon201pfem}, and both implementations can be found in the associated codebase.}
We use $N_p=100$ particles for filtering, and sample $N_s = 10$ smoothed trajectories using FFBSi.

\begin{figure}[t]
    \centering
    \includegraphics[width=\columnwidth]{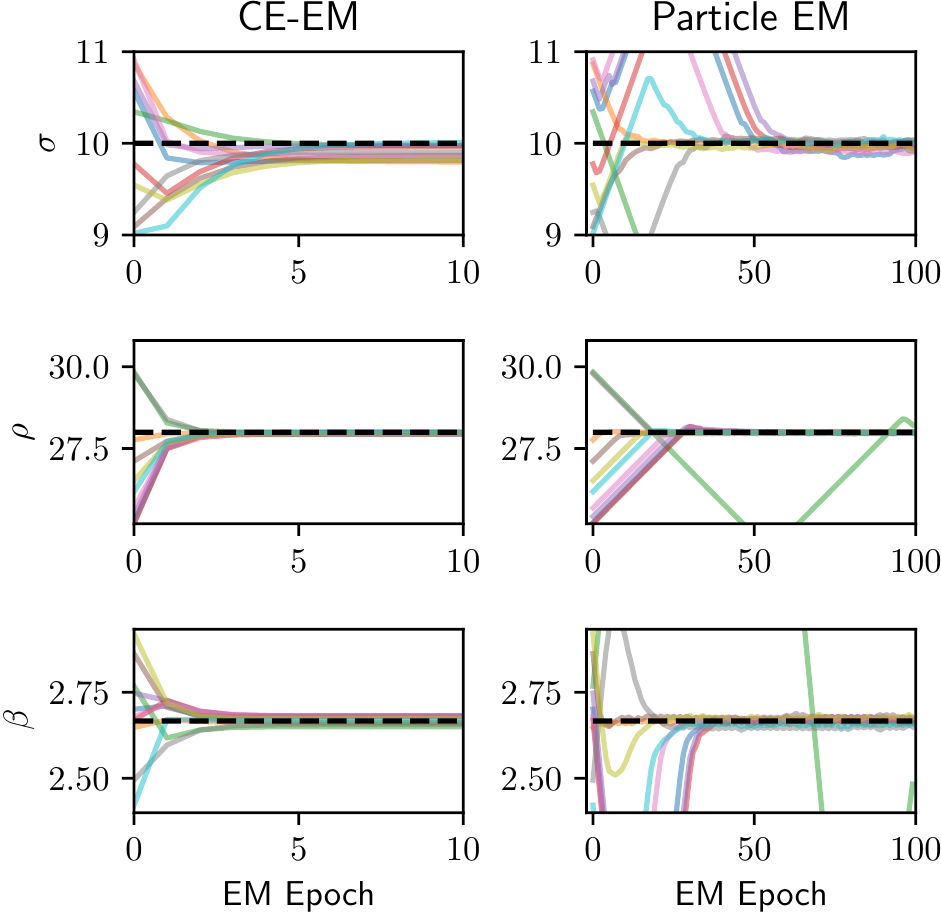}
    \caption{Comparison of CE-EM and Particle EM on parameter estimation of a single Lorenz system.}
    \label{fig:particleemcomp}
\end{figure}

\Cref{fig:particleemcomp} shows the estimated parameters versus EM epoch using CE-EM and Particle EM. 
We plot learning curves for 10 random seeds, each of which initializes parameter estimates to within 10\% of their true value, and uses a different set of training trajectories.
We see that CE-EM consistently converges to accurate parameter estimates in approximately 5 epochs. 
Estimates of Particle EM appear to initially diverge but in all but one case converge to a similar accuracy in 50 epochs.
Furthermore, since the complexity of FFBSi is $O(N_p N_s)$,\footnote{Variants of FFBSi algorithm improve on the $O(N_p N_s)$ complexity \cite{lindsten2013backward}.} the runtime per epoch of Particle EM is $47\times$ more than that of CE-EM.\footnote{Experiments were run on a computer with an Intel® Core™ i7-6700K CPU @ 4.00GHz $\times 8$ processor and 32GB of memory. Operations are parallelized across trajectories for both implementations.}
Since the variance of particle-based methods generally increases with the effective dimension of the system, the bias induced by CE-EM may be a worthwhile trade-off for fast and reliable parameter estimation.

\subsubsection{Convergence of CE-EM on High-Dimensional Problems}
To demonstrate that CE-EM is capable of identifying high-dimensional systems, we show that we can estimate the dynamics of an $18$ dimensional system of six coupled Lorenz attractors.
Moreover, we test whether CE-EM converges to more accurate estimates when more trajectories are provided. 
To test these claims, we sample $2$, $4$, and $8$ trajectories from a deterministic system with parameters $\theta_\text{true}$, and $\sigma_v=0.01$.
We randomly initialize each element of the parameters being optimized ($\theta = [\sigma_{1:K}, \rho_{1:K}, \beta_{1:K}, H]$) to within 10\% of the their value in $\theta_\text{true}$.
We then run CE-EM on each batch, tracking the error in the estimated dynamics as training proceeds. 
We measure this error, which we call $\epsilon(\theta)$, as follows:
\begin{equation}
    \epsilon(\theta)=\mathbf{E}_{x\sim p(x_0)}\left[\lVert f_{\theta}(x) - f_{\theta_\text{true}}(x)\rVert_2\right]
\end{equation}
In the learning step, we do not regularize $\theta$ to a prior and set $\rho_\theta=0$.

For comparison, we also run the same Particle EM implementation as in the previous experiment on this problem. 
We use the same $N_p, N_s$, and hyperparameters as before. 

\begin{figure}[t]
     \includegraphics[width=1.0\columnwidth]{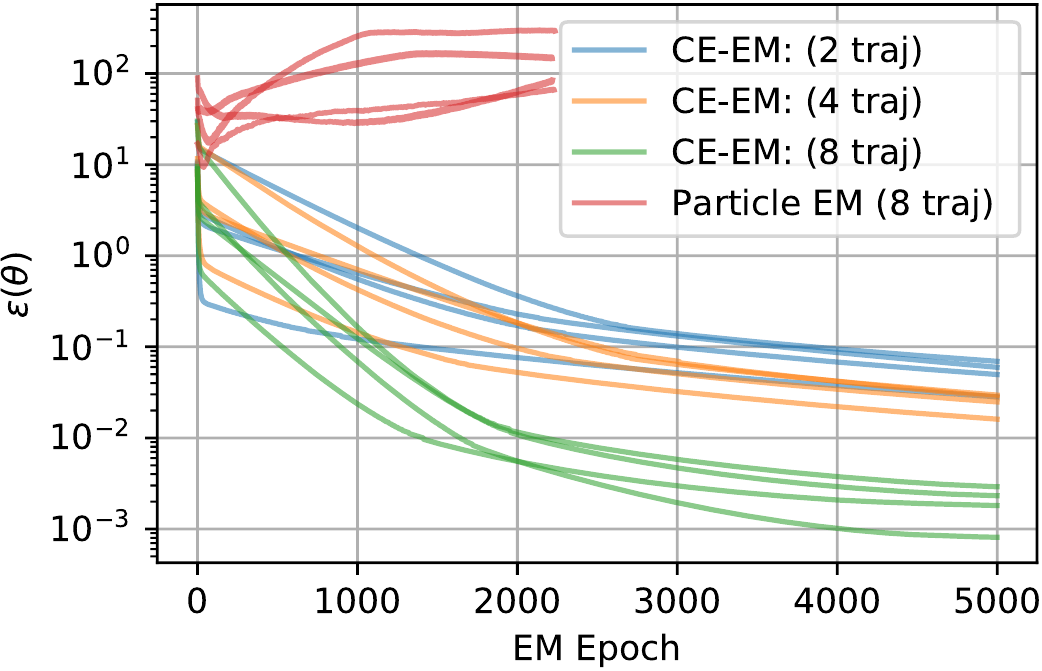}
\vspace{\figtocap}
\caption{Error in estimated dynamics as CE-EM trains on a varied number of trajectories from a system of six coupled Lorenz attractors.} \label{fig:lorenz_conv}
\end{figure}

\Cref{fig:lorenz_conv} shows the results of this experiment for four random seeds for each batch size. 
We can see that, as the number of trajectories used in training increases, the error in the estimated dynamics tends toward zero. 
Furthermore, we see that CE-EM convergences monotonically to a local optimum in all cases. 
In contrast, Particle EM appears to initially improve but then converges to very poor parameter estimates. 

The experiments conducted thus far have demonstrated that CE-EM can learn unbiased parameter estimates of nearly-deterministic systems, and can scale to high-dimensional problems for which particle-based methods are intractable. 
In the next experiment, we use CE-EM to characterize the effect of unobserved states on the dynamics of an aerobatic helicopter.

\subsection{Characterizing Aerobatic Helicopter Dynamics}
\label{sec:heli}

Characterizing the dynamics of a helicopter undergoing aggressive aerobatic maneuvers is widely considered to be a challenging system-identification problem~\citep{abbeel2010heli,punjani2015deep}. 
The primary challenge is that the forces on the helicopter depend on the induced state of the fluid around it. The state of the fluid cannot be directly observed and its dynamics model is unknown. Merely knowing the state of the helicopter and the control commands at a given time does not contain enough information to accurately predict the forces that act on it.

In order to address this issue, \cite{punjani2015deep} use an approach based on Takens theorem, which suggests that a system's state can be reconstructed with a finite number of lagged-observations of it~\citep{takens1981detecting}. 
Instead of attempting to estimate the unobserved fluid state, they directly learn a mapping from a $0.5\,\si{\second}$ history of observed state measurements and control commands to the forces acting on the helicopter. 

This approach is equivalent to considering the past $0.5\,\si{\second}$ of observations as the system's state. However, it can require a very large number of lagged observations to represent complex phenomena. In reality, the characteristic time of unsteady flows around helicopters can easily be up to tens of seconds.
Having such a high dimensional state can make the control design and state-estimation more complicated. 
To avoid large input dimensions, a trade-off between the duration of the history and sample frequency is necessary. 
This trade-off will either hurt the resolution of low-frequency content or will alias high-frequencies. 
We attempt to instead explicitly model the unobserved states affecting the system.

\begin{figure*}[t]
    \centering
    \begin{subfigure}[b]{0.49\columnwidth}
        \includegraphics[width=\columnwidth]{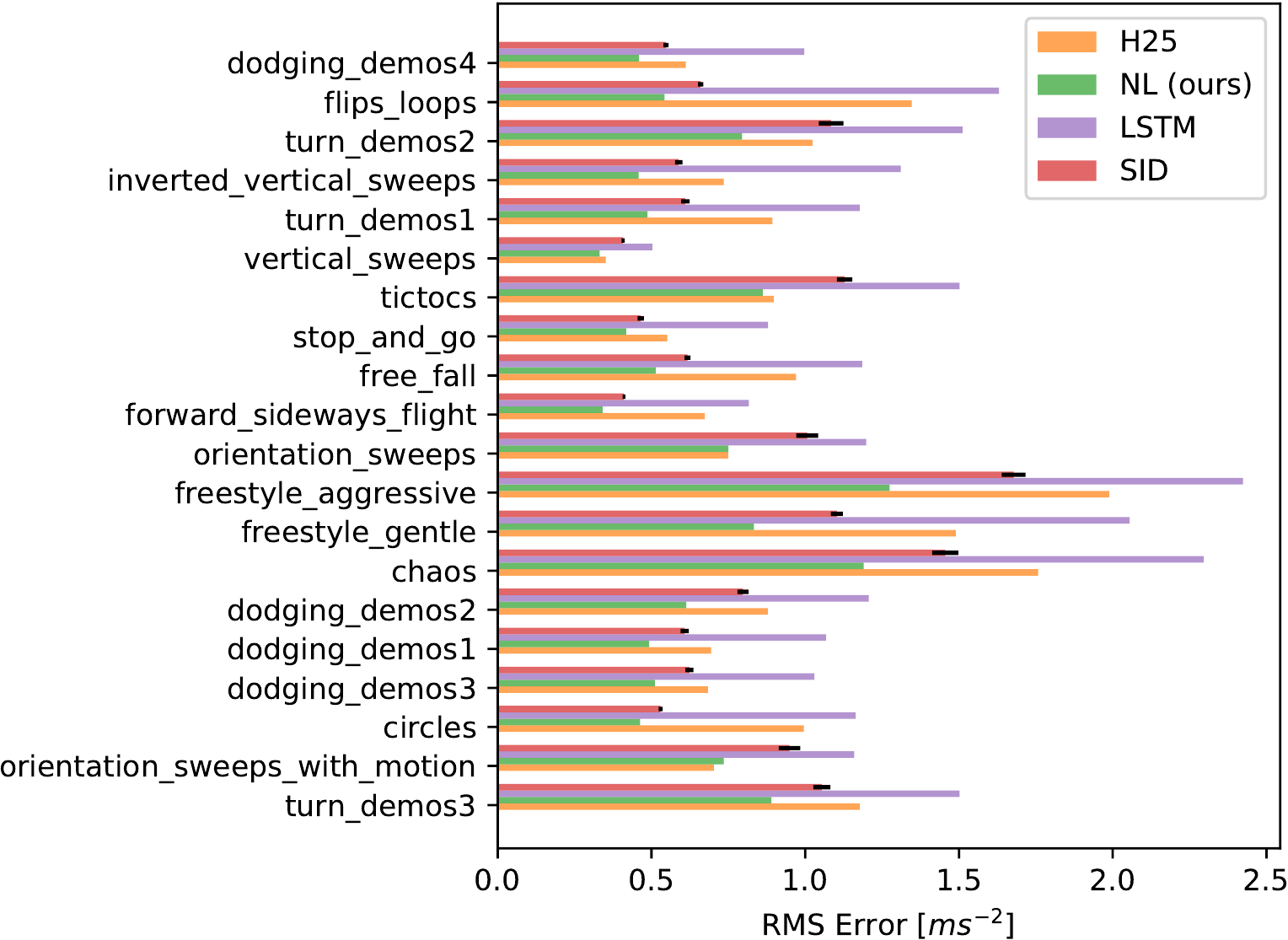}
        \caption{}
        \label{fig:testperf}
    \end{subfigure}
    \begin{subfigure}[b]{0.49\columnwidth}
        \centering
        \includegraphics[width=\columnwidth]{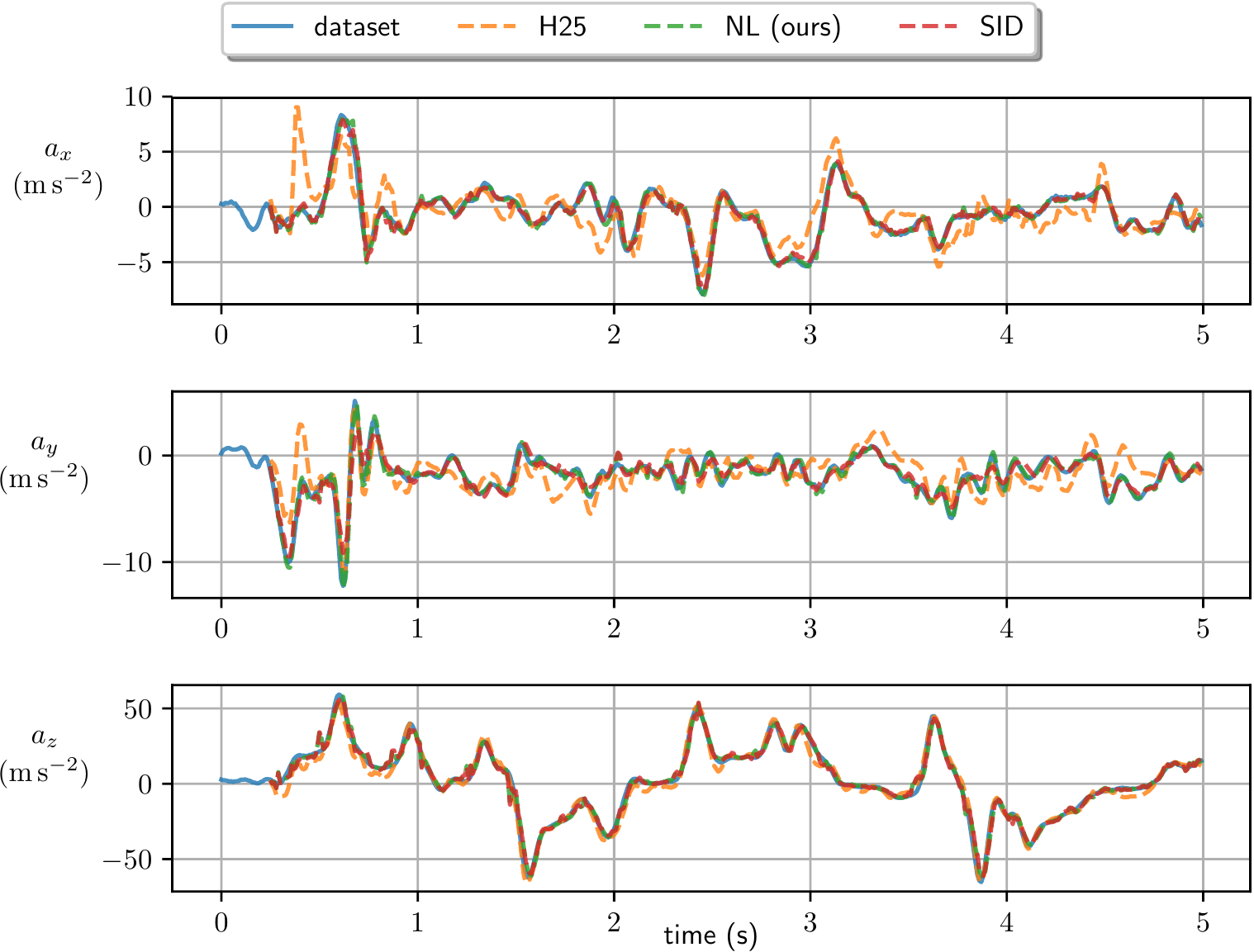}
        \caption{}
        \label{fig:accelerr}
    \end{subfigure}
    \caption{(a): Test performance of optimized models on various trajectories. Error bars on \textit{SID} represent the standard deviation of performance of the 10 trained models. (b): Predicted acceleration along axis $x$, $y$ and $z$ in the body frame for one of the harder test set trajectories. %
    Larger versions of these plots, including rotational accelerations, can be found in \Cref{app:large_fig}.}
\end{figure*}

\subsubsection{Objective and dataset}

The objective of this learning problem is to predict $y_t$, the helicopter's acceleration at time $t$, from an input vector $u_t$ containing the current measured state of the helicopter (its velocity and rotation rates) and the control commands. 

We use data collected by the Stanford Autonomous Helicopter Project~\citep{abbeel2010heli}. Trajectories are split into $10\,\si{\second}$ long chunks and then randomly distributed into train, test, and validation sets according to the established protocol \cite{abbeel2010heli, punjani2015deep} and summarized in \Cref{app:helidata_protocol}. The train, test, and validation sets respectively contain 466, 100, and 101 trajectories of 500 time-steps each. 

A simple success metric on a given trajectory is the root mean squared prediction error,
\begin{equation}
\text{RMSE} =  \sqrt{\frac{1}{T} \sum_{t=1}^T \left\lVert y_t^\text{(measured)} - y_t^\text{(pred)} \right\rVert^2_2} 
\end{equation}
where $y_t^\text{(measured)}$ is the measured force from the dataset, $y_t^\text{(pred)}$ is the force predicted by the model, and $T$ is the number of time-steps in each trajectory.

\subsubsection{Previous work and baselines}

\textit{Naive:} We first consider a naive baseline that does not attempt to account for the time-varying nature of the fluid-state. We train a neural-network to map only the current helicopter state and control commands to the accelerations: $y_t = \text{NN}_{\theta_n}(u_t)$, where $\text{NN}_{\theta_n}$ is a neural-network with parameters $\theta_n$.
We refer to this model as the \textit{naive model}. 

\textit{H25:} We also compare to the work of \citet{punjani2015deep}. 
They predict $y_t$ using a time-history $u_{t-H:t}$ of $H=25$ lagged observations of the helicopter's measured state and control commands. This input is passed through a ReLU-activated neural network with a single hidden-layer combined with what they call a Quadratic Lag Model. 
As a baseline, we reproduce their performance with a single deep neural network $y_t = \text{NN}_{\theta_h}(u_{t-H:t})$ with parameters $\theta_h$.
We call this neural network model the \textit{H25 model}. 
Both of these models can be trained via stochastic gradient descent to minimize the Mean-Squared-Error (MSE) of their predictions for $y$. 
The optimization methodology for these models is described in \Cref{app:heliopt}.

\textit{SID:} As a third baseline, we compare with subspace-identification methods~\citep{van2012subspace}. 
We let $\tilde{y}_t = y_t - \text{NN}_{\theta_n}(u_t)$ be the prediction errors of the trained naive model. 
We use the MATLAB command \texttt{n4sid} to fit a linear dynamical system of the following form:
\def\S{\text{s}}
\begin{equation}
    \begin{aligned}
        x_{t+1} &= A_\S x_t + B_\S u_t; &\quad \tilde{y}_t &= C_\S x_t + D_\S u_t
    \end{aligned}
\end{equation}
Here, $x\in R^d$ is the unobserved state with arbitrary dimension $d$. The learned parameters are $\theta_\S = [A_\S,B_\S,C_\S,D_\S]$. 
We use a state dimension of $10$ and call this model the \textit{SID model}.
The \texttt{n4sid} algorithm scales super-linearly with the amount of data supplied, and thus we train on $10$ randomly sampled subsets of 100 trajectories each, and report the distribution in prediction performance. This approach fits a linear system to the \textit{residual error} of the naive model, therefore the prediction of $y_t$ obtained from the \textit{SID model} is a nonlinear function of the states $x_t$.

\textit{LSTM:} We also train an LSTM~\cite{hochreiter1997lstm} on the residual time-series $\tilde{y}_{1:T}$ and $u_{1:T}$.

\subsubsection{Non-linear Unobserved State Model}
We use CE-EM to train a non-linear SSM.
Similar to the parameterization used for subpace-identification, we fit the {prediction errors} of the naive model using the following dynamical system:
\def\NL{ \text{\tiny NL}}
\begin{equation}
    \begin{aligned}
        x_{t+1} &= A_\NL x_t + B_\NL u_t  \\ \tilde{y}_t &= C_\NL x_t + D_\NL u_t + \text{NN}_{\bar{\theta}_{\NL}}(x_t,u_t)
    \end{aligned}
\end{equation}
where $\text{NN}_{\bar{\theta}_\NL}$ is a neural network, and $\theta_\NL =[A_\NL,B_\NL,C_\NL,D_\NL,\bar{\theta}_\NL]$ are the learned parameters.
We introduce non-linearity only in the observation function because it is known from Koopman theory that a non-linear system can be approximated by a high-dimensional linear system provided the correct non-linear mapping between them~\cite{brunton2016koopman}.

While learning, we assume that both process and observation noise are distributed with diagonal Gaussian covariance matrices $\sigma_w\mathbf{I}$ and $\sigma_v\mathbf{I}$ respectively. 
The values of $\sigma_w$ and $\sigma_v$ are treated as hyperparmeters of CE-EM, and are both set to 1.
Here as well, we use a state dimension of $10$ and call this model the \textit{NL model}. 
The optimization methodology for this model is described in \Cref{app:heliopt}.

It should be noted that the system we learn need not actually correspond to an interpretable model of the fluid-state, but only of time-varying hidden-states that are useful for predicting the accelerations of the helicopter.
Expert knowledge of helicopter aerodynamics could be used to further inform a gray-box model trained with CE-EM.

\subsubsection{Evaluation methodology}

The test RMSE of the \textit{naive}, \textit{H25}, and \textit{LSTM} models can be evaluated directly on the test trajectories using next-step prediction. However, the \textit{SID} and \textit{NL} models require an estimate of the unobserved state before making a prediction. 
The natural analog of next-step prediction is extended Kalman filtering (EKF), during which states are recursively predicted and corrected given observations. 
At a given time-step, a prediction of $\tilde{y}_t$ is made using the current estimate of $x_t$, and is used in the computation of RMSE.
The state-estimate is then corrected with the measured $\tilde{y}_t$.

\subsubsection{Results}

\Cref{fig:testperf} shows the RMSE of the compared models on trajectories in the test-set. 
We see that the \textit{NL} model is able to consistently predict the accelerations on the helicopter with better accuracy than any of the other models.
The naive model performs on average 2.9 times worse than the \textit{H25} model, and its results can be found in \Cref{app:large_fig}.
The \textit{LSTM} model also performs poorly, on average 1.4 times worse than the \textit{H25} model.
The \textit{SID} model notably outperforms the state-of-the-art \textit{H25} model, suggesting that a large linear dynamical system can be used to approximate a non-linear and partially observable system~\citep{Korda2018}.
However, introducing non-linearity as in the \textit{NL} model noticeably improves performance.

\Cref{fig:accelerr} depicts the errors in prediction over a sample trajectory in the test-set. 
Here, we also see that the \textit{NL} model is able to attenuate the time-varying error present in predictions made by the \textit{H25}, suggesting that it has accurately characterized the dynamics of unobserved, time-varying states.

This experiment validates the effectiveness of CE-EM to identify a non-linear dynamical model of unobserved states that affect the forces acting an aerobatic helicopter.

\section{Conclusions}
\label{sec:conclusion}

This paper presented an algorithm for system identification of non-linear systems given partial state observations.
The algorithm optimizes system parameters given a time history of observations by iteratively finding the most likely state-history, and then using it to optimize the system parameters.
The approach is particularly well suited for high-dimensional and nearly deterministic problems.

In simulated experiments on a partially observed system of coupled Lorenz attractors, we showed that CE-EM can perform identification on a problem that particle-based EM methods are ill-suited for. However, we also find that CE-EM yields biased parameter estimates in the presence of large process noise.
This bias can be partially mitigated by locally approximating the posterior state-marginals as Gaussian, as is done by MAP-EM~\cite{goodwin2005approximate}.
We then used the algorithm to model the time-varying hidden-states that affect the dynamics of an aerobatic helicopter.
The model trained with CE-EM outperforms state-of-the-art methods because it is able to fit large non-linear models to unobserved states.

Numerous system-identification problems can be studied using CE-EM.
Recently, there have been tremendous efforts to characterize predictive models for the spread of COVID-19~\cite{fanelli2020epianalysis}. 
Limited capacity for testing the prevalence of the disease makes relevant states partially observed, and thus CE-EM may be useful for its modeling.
We also hope to apply CE-EM to very high-dimensional systems with sparsely coupled dynamics using general-form consensus optimization~\cite{boyd2011admm}.

\section*{Acknowledgments}
This work is supported in part by DARPA
under agreement number D17AP00032, and by the King Abdulaziz City for Science and Technology (KACST) through the Center of Excellence in Aeronautics and Astronautics. The content is solely the responsibility of the
authors and does not necessarily represent the official views of DARPA or KACST.

\bibliography{references}
\bibliographystyle{icml2020}

\clearpage
\appendix
\section{Appendix}
\subsection{Dataset Collection and Preprocessing}\label{app:helidata_protocol}

In this work we use the dataset gathered by \citet{abbeel2010heli} and available at \url{http://heli.stanford.edu/}. A gas-powered helicopter was flown by a professional pilot to collect a large dataset of $6290\si{\second}$ of flight. There are four controls: the longitudinal and lateral cyclic pitch, the tail rotor pitch and the collective pitch. The state is measured thanks to an accelerometer, a gyroscope, a magnetometer and vision cameras. \citet{abbeel2010heli} provide the raw data, as well as states estimates in the Earth reference frame obtained with extended Kalman smoothing. Following \citet{punjani2015deep}, we use the fused sensor data and downsample it from 100Hz to 50Hz.  

From the Earth frame accelerations provided in the dataset, we compute body frame accelerations (minus gyroscopic terms) which are the prediction targets for our training.
Using the notations from \citet{punjani2015deep}, we can write the helicopter dynamics in the following form:
\begin{equation}
    \dot{s} = F(s,\delta) = \begin{bmatrix}C_{12}v\\\frac12\hat{\omega}q\\C_{12}^\top g - \omega\times v + f_v(s,\delta)\\f_\omega(s,\delta) \end{bmatrix}
\end{equation}
where $s\in\bbR^{13}$ is the helicopter state consisting of its position $r$, quaternion-attitude $q$, linear velocity $v$, angular velocity $\omega$, and $\delta \in \bbR^4$ to be the control command. 
$C_{12}$ is the rotation-matrix from the body to Earth reference frame, and $f_v$ and $f_\omega$ are the linear and angular accelerations caused by aerodynamic forces, and are what we aim to predict. 

The above notation is related to that used in \Cref{sec:heli} as follows:

\begin{itemize}
    \item We define $u$ as the concatenation of all inputs to the model, including the relevant state variables $v$ and $\omega$ and control commands $\delta$.
    \item We define $y$ as the output predicted, which would correspond to a concatenation of  $f_v$ and $f_\omega$.
    \item We define $x$ as the vector of unobserved flow states to be estimated and is not present in their model.
\end{itemize}

The processed dataset used in our experiments can be found at \url{https://doi.org/10.5281/zenodo.3662987}~\cite{procdataset}.

\subsection{Training Helicopter Models}\label{app:heliopt}

Neural networks in the \textit{naive} and \textit{H25} models have eight hidden layers of size $32$ each, and tanh non-linearities.
We optimize these models using an Adam optimizer~\citep{kingma2014adam} with a harmonic learning rate decay, and mini-batch size of $512$. 

The neural network in the \textit{NL} model has two hidden layers of size $32$ each, and tanh non-linearity.
We train the \textit{NL} model with CE-EM, using $\rho_x=\rho_\theta=0.5$, $\sigma_w=\sigma_v=1.0$, and use an Adam optimizer to optimize \Cref{eqn:learning} in the learning step. 
The learning rate for dynamics parameters in $\theta_{\NL}$ is $5.0\times 10^{-4}$ and observation parameters in $\theta_{\NL}$ is $1.0 \times 10^{-3}$.
For its relative robustness, we optimize \Cref{eqn:smoothinguncon} using a non-linear least squares optimizer with a Trust-Region Reflective algorithm~\citep{scipy} in the smoothing step. This step can be solved very efficiently by providing the solver with the block diagonal sparsity pattern of the Jacobian matrix.

\label{app:ekf}
To evaluate the test metric, running an EKF is required. The output of an EKF depends on several user-provided parameters:
\begin{itemize}
    \item $x_0$: value of the initial state
    \item $\Sigma_0$: covariance of error on initial state
    \item $Q$: covariance of process noise
    \item $R$: covariance of observation noise
\end{itemize}
In this work, we assume that $Q$, $R$ and $\Sigma_0$ are all set to the identity matrix. $x_0$ is assumed to be 0 on all dimensions. 

A well-tuned EKF with an inaccurate initial state value converges to accurate estimations in only a few time steps of transient behavior. 
Since the \textit{H25} model needs 25 past inputs to predict its first output prediction, we drop the first 25 predictions from the EKF when computing RMSE, thereby omitting some of the transient regime.

\subsection{Figures}\label{app:large_fig}

\begin{figure*}[h]
    \centering
    \includegraphics[width=\textwidth,height=\textheight,keepaspectratio]{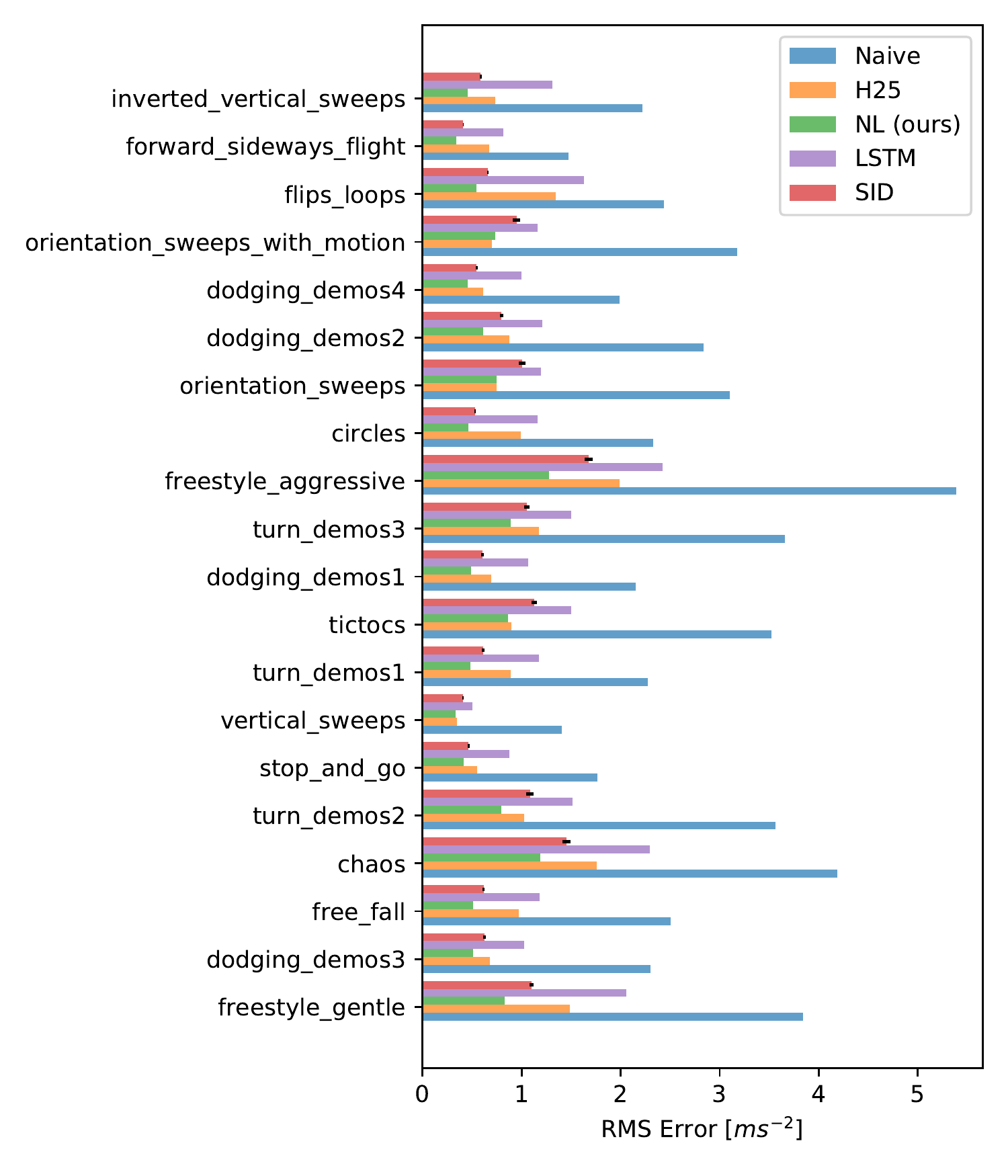}
    \caption{Test performance of optimized models on various trajectories}
\end{figure*}

\begin{sidewaysfigure}[h]
     \includegraphics[width=\textwidth,height=\textheight,keepaspectratio]{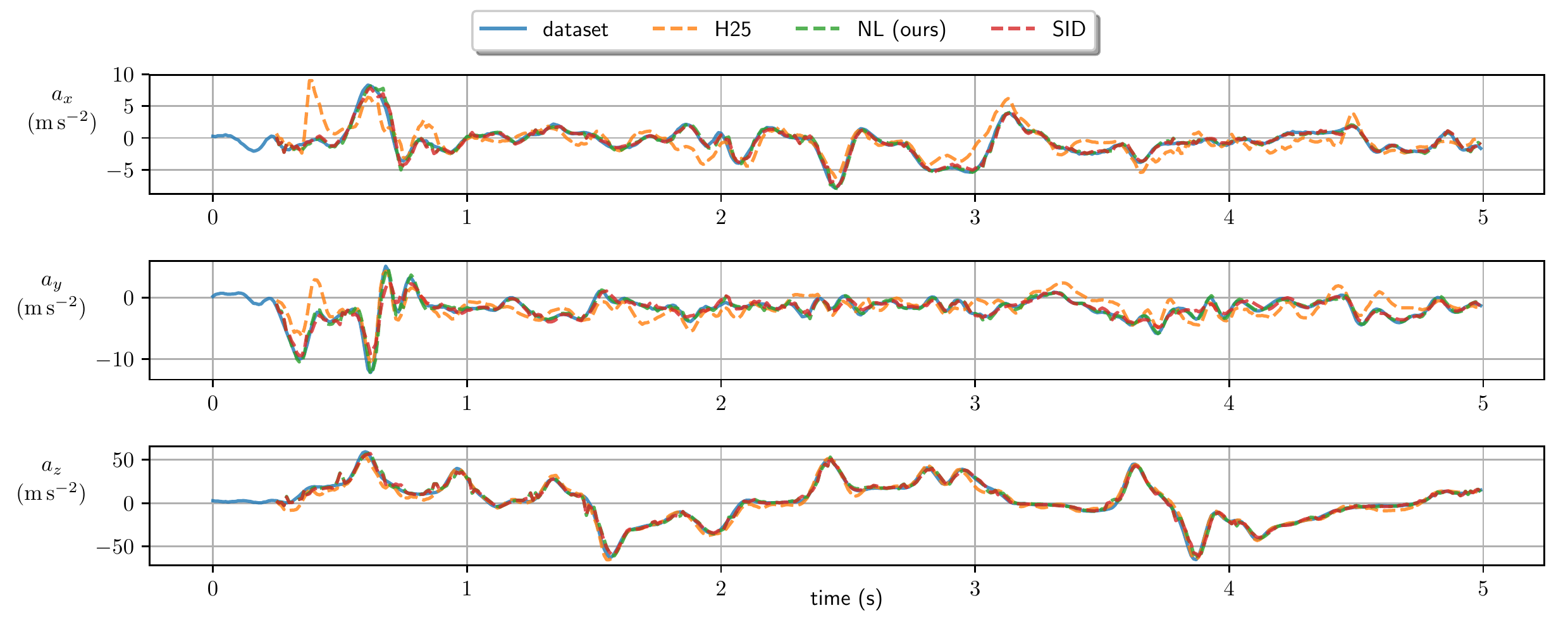}
\caption{Predicted acceleration along axis $x$, $y$ and $z$ in the body frame. For subspace identification and models trained by CE-EM, this plot requires running an extended Kalman filter. These figures can be reproduced for any other trajectory with the included code.}\label{fig:large_traj_acc}
\end{sidewaysfigure}

\begin{sidewaysfigure}[h]
     \includegraphics[width=\textwidth,height=\textheight,keepaspectratio]{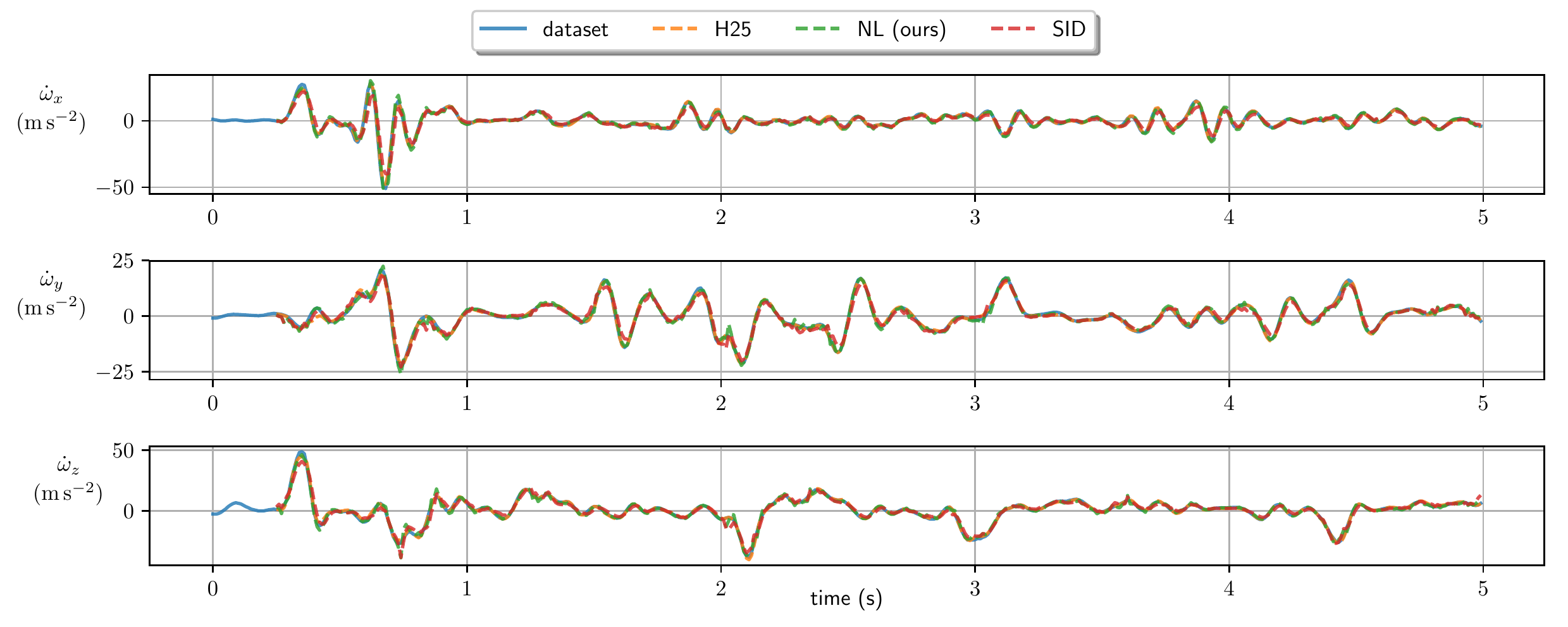}
\caption{Predicted circular accelerations around axis $x$, $y$ and $z$ in the body frame. For subspace identification and models trained by CE-EM, this plot requires running an extended Kalman filter. These figures can be reproduced for any other trajectory with the included code.}\label{fig:large_traj_mom}
\end{sidewaysfigure}

\end{document}